\documentclass[letterpaper, 10 pt, conference]{ieeeconf}
\IEEEoverridecommandlockouts
\usepackage{cite}
\usepackage{amsmath,amssymb,amsfonts}
\usepackage{algorithm}
\usepackage{algpseudocode}
\usepackage{graphicx}
\usepackage{textcomp}
\usepackage{xcolor}
\usepackage{subcaption}  
\usepackage{tabularx}

\def\BibTeX{{\rm B\kern-.05em{\sc i\kern-.025em b}\kern-.08em
    T\kern-.1667em\lower.7ex\hbox{E}\kern-.125emX}}

\usepackage{xparse}

\title{Trajectory Planning Using Safe Ellipsoidal Corridors as \\Projections of Orthogonal Trust Regions}
\author{Akshay Jaitly$^{1}$, Jon Arrizabalaga$^{2}$, and Guanrui Li$^{1}$
\thanks{$^{1}$A. Jaitly and G. Li are with the Robotics Engineering Department, Worcester
Polytechnic Institute, Worcester, MA 01609, USA {\tt\small \{ajaitly, gli7\}@wpi.edu}}
\thanks{$^{2}$ J. Arrizabalaga is with the Robotics Institute at Carnegie Mellon University {\tt\small jarrizab@andrew.cmu.edu}}
\thanks{We would like to thank Ethan Chandler regarding discussions about the acceleration of Trust Region methods using Eigendecomposition and thank Arvind Raghunathan for discussions about the dual of QCQPs.}}
\begin{document}
\maketitle

\begin{abstract}
Planning collision free trajectories in complex environments remains a core challenge in robotics. Existing corridor based planners which rely on decomposition of the free space into collision free subsets scale poorly with environmental complexity and require explicit allocations of time windows to trajectory segments. We introduce a new trajectory parameterization that represents trajectories in a nonconvex collision free corridor as being in a convex cartesian product of balls. This parameterization allows us to decouple problem size from geometric complexity of the solution and naturally avoids explicit time allocation by allowing trajectories to evolve continuously inside ellipsoidal corridors. Building on this representation, we formulate the Orthogonal Trust Region Problem (Orth-TRP), a specialized convex program with separable block constraints, and develop a solver that exploits this parallel structure and the unique structure of each parallel subproblem for efficient optimization. Experiments on a quadrotor trajectory planning benchmark show that our approach produces smoother trajectories and lower runtimes than state-of-the-art corridor based planners, especially in highly complicated environments.
\end{abstract}

\section{ Introduction }
Collision free trajectory optimization is a core challenge in robotics, useful in mobile robot navigation and mobile manipulation~\cite{marcucci2023gcs, PallarLiLoianno2025_CBF_Manip_ICRA, MaoEtAl2021_PersepPerching_IROS}. By mathematically defining sets of feasible trajectories that a system can undergo, and assigning a cost to each element of the set, trajectory optimization can be posed as a mathematical programming problem. Several methods exist to characterize these sets to facilitate efficient searches for optimal feasible trajectories.

Conventional sampling based planners~\cite{LaValle1998RRT, karamanSamplingbasedAlgorithmsOptimal2011, kavraki1996probabilistic} build tree or graph like structures in the configuration space, with nodes corresponding to collision free states and edges representing locally feasible motions. Paths through these structures define potential trajectories, which can then be searched efficiently using graph search or shortest path algorithms. On the other hand, some optimization based approaches~\cite{TracyHowellManchester2023, foehnTimeoptimalPlanningQuadrotor2021, LiLoianno2023_NMPC_CoopTransp_IROS, thirugnanamSafetyCriticalControlPlanning2022,GoarinLiSavioloLoianno2025_DNMPC_Safety_ICRA, JaitlyMERL2025} impose constraints on robot’s states (enforcing that a barrier between collision objects is respected) to define sets of feasible solutions, then minimize a cost function within this nonconvex feasible set. These enable powerful nonconvex optimization solvers for trajectory optimization. In more complicated scenarios, with larger dimensional spaces or with numerous obstacles, the above methods inevitably face the curse of dimensionality.

Recent work has introduced convex approximations of free space through “safe corridors”, using either polytopic or ellipsoidal regions \cite{deitsComputingLargeConvex2015,PetersenTedrake2023IRISNP,liuPlanningDynamicallyFeasible2017,renBubblePlannerPlanning2022}. In these approaches, a trajectory is divided into segments, and each segment is constrained to lie within a corresponding convex subset of the obstacle free space at a specific point in time. By enforcing convex constraints on the robot’s configuration, these methods convert a difficult global planning problem into convex optimization, which can be solved using efficient convex programming techniques \cite{marcucci2023gcs,liuPlanningDynamicallyFeasible2017,renBubblePlannerPlanning2022}.

However, a key challenge in these methods is allocating time across the convex regions. Each region requires a specified time interval, and if this interval is too short, the system must accelerate sharply to meet continuity constraints, while overly long intervals produce inefficient motions. As the number of convex subsets increases, these timing decisions become increasingly difficult to manage, often introducing sensitivity and limiting the scalability of convex corridor methods.

\begin{figure}[t]
    \centering  
    \includegraphics [width=\columnwidth,  trim=60 100 40 125, clip] {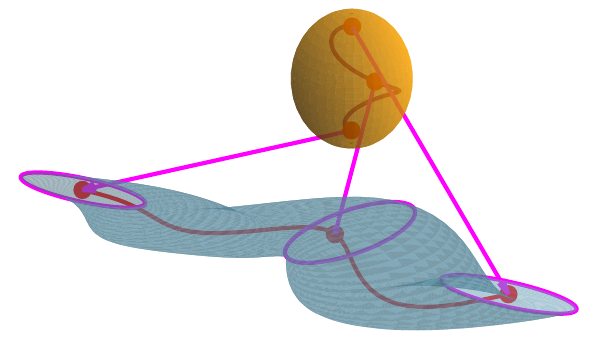}
    \caption{A path (red) within an ellipsoidal corridor (blue) as a time varying projection of a \textbf{trajectory} within a lifted hypersphere (orange). Note that projected points are  restricted to be within the relevant ellipsoidal cross section.}\label{fig:proj_traj_to_corridor}
\end{figure}

\cite{arrizabalagaDifferentiableCollisionFreeParametric2024a} introduced Differentiable Collision Free Parametric Corridors which model free space as a smooth nonconvex corridor, made of a continuously deforming convex set rather than discrete segments, offering a more unified description of collision free regions. Building on this idea, we view these nonconvex corridors as time-varying projections of orthogonal trust regions in a higher dimensional parameter space. By lifting trajectories into a space where feasible solutions form a cartesian product of high dimensional balls, each point in this lifted space naturally corresponds to a collision free trajectory within the corridor. This decouples problem size from environmental complexity while allowing representations of paths in the safe corridor as points in the lifted space.

This representation results in a favorable convex feasible set for trajectory optimization. We pose the resulting problem as the Orthogonal Trust Region Problem (Orth-TRP), which can be expressed as a collection of interconnected trust region subproblems (TRPs). Because each block of variables has its own seperable constraint, the Orth-TRP naturally supports a parallelizable algorithm, where each block can be updated efficiently using trust region steps that resemble simple one dimensional line searches.

In summary, our contributions are:

\begin{itemize}
    \item We develop a convex representation of sets of trajectories within a nonconvex set of configurations using a product of multiple high dimensional balls, decoupling the solution complexity from the size of the optimization problem.

    \item We create a solver that exploits both, the parallelizable structure and the unique Trust-Region-like structure of our resulting problem to solve quadratically constrained quadratic optimization problems where constraints are enforced on separable variables.
\end{itemize}


Taken together, these components create a scalable and geometrically intuitive framework for collision-free trajectory optimization in complex environments. Our experiments show that this approach consistently outperforms state-of-the-art implementations on challenging benchmarks. In direct comparisons with the method of \cite{liuPlanningDynamicallyFeasible2017}, even when using a powerful solver like OSQP \cite{StellatoBanjacGoulartBemporadBoyd2020_OSQP}, our method solves long-horizon planning problems faster while producing comparable or smoother results. To the best of our knowledge, this is the first continuous parameterization of trajectories that enables optimization within a collision-free corridor where problem size and runtime are agnostic to horizon length and solution complexity. We achieve this by combining a new convex programming formulation with a solver that fully exploits the problem’s separable trust-region structure, eliminating the scaling and time-allocation issues inherent to existing corridor-based approaches and enabling both speed and robustness at scale.

\section{ Background }

In this work, we develop a parameterization of paths within Parametric Collision-Free Corridors. These are time-varying convex (ellipsoidal) sets embedded in the obstacle-free space. By construction, any path inside these corridors is guaranteed to be collision free. We formulate trajectory optimization within this corridor as minimizing a quadratic objective over a ball-constrained set, matching the structure of the classical Trust Region Problem. Accordingly, this section reviews key concepts from Parametric Collision Free Corridors and the classical Trust Region Problem, which form the basis of our approach. The notation used throughout is summarized in Table~\ref{tab:notation}.

\begin{table}[t]
\centering
\caption{Notation}
\label{tab:notation}
\renewcommand{\arraystretch}{1.3} 
\begin{tabularx}{\linewidth}{>{\raggedright\arraybackslash}m{0.25\linewidth}X}
\hline
\textbf{Symbol} & \textbf{Meaning} \\
\hline
$q \in \mathbb{R}^{N_q}$ & Configuration in the configuration space. \\
\hline
$\epsilon \in [0,1]$ & Central “time” variable along the trajectory; $\epsilon=1$ denotes the end of the trajectory. \\
\hline
$y \in \mathbb{R}^{N_y}$ & Vector in the lifted parameter space, with $N_y \geq N_q$. A configuration space trajectory is given by $q(t)=L(t)y+\tilde{q}(t)$. \\
\hline
$\mathcal{S}^{N_a}$ & Unit ball in $\mathbb{R}^{N_a}$, defined as 
$\mathcal{S}^{N_a}=\{a\in\mathbb{R}^{N_a}:\|a\|\le 1\}$ 
(in contrast to using $\|a\|=1$ to describe only the surface). \\
\hline
$\mathcal{A} = B\mathcal{C}$ & For set valued $\mathcal{A},\mathcal{C}$ and matrix valued $B$, represents 
$\mathcal{A}=\{a=Bc\mid c\in\mathcal{C}\}$. \\
\hline
$F(t)$ and $F_t$ & Used interchangeably to refer to some function evaluated at $t$. \\
\hline\\
\end{tabularx}
\end{table}

\subsection{Convex Ellipsoidal Corridors} \label{sec:background_ellipsoidal_corridor}

\cite{arrizabalagaDifferentiableCollisionFreeParametric2024a} proposed a framework for constructing parametric collision free corridors in which each cross section is an ellipsoid whose center, orientation, and shape vary smoothly with a path parameter $\epsilon$. By expressing these parameters as low degree polynomials of $\epsilon$, they obtain a differentiable corridor that follows the reference path and maximizes free space volume while remaining obstacle free. At each $\epsilon$ the cross section is restricted to the plane orthogonal to the reference trajectory and can be written as
\[
\mathcal{C}_\epsilon=
\Bigl\{
q=r_\epsilon+ Tq_\perp \;\big|\; 
q_\perp^\top E_\epsilon q_\perp + d_\epsilon^\top q_\perp \le1,\;
q_\perp\in\mathbb{R}^{N_q-1}
\Bigr\},
\]
where $T$ defines the axes that span the plane perpendicular to $\dot r_\epsilon$. $E_\epsilon\succ0$, $d_\epsilon$ define the ellipsoid. The corridor parameters are obtained by solving a linear program that maximizes corridor size subject to collision avoidance constraints.

\subsection{Solving the Trust Region Problem}
The classical Trust Region Problem (TRP) minimizes a quadratic objective subject to a norm bound
\begin{equation}
\begin{split}
\min_{x} &\,\,\, \tfrac{1}{2}x^\top Qx + g^\top x
\\
s.t.&\,\,\, \|x\| \leq \Delta.
\end{split}
\end{equation}

\begin{figure}
    \centering  \includegraphics[width=0.45\columnwidth] {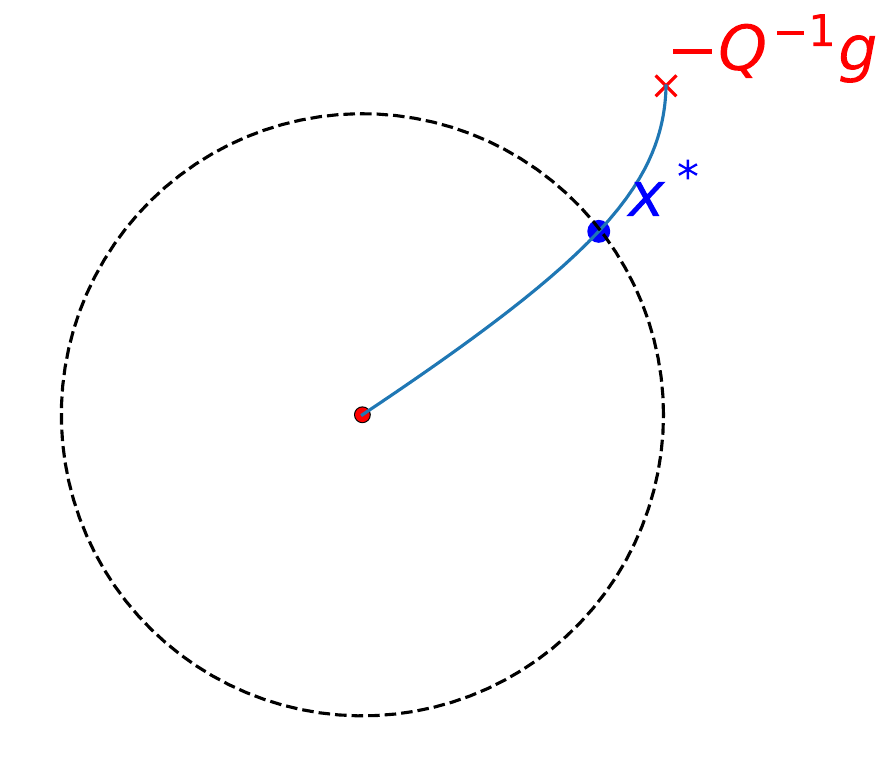}
    \caption{The optimal curve in the Trust Region Problem. Each point on the curve respects the "stationarity conditions" of optimality.}\label{fig:TRP_opt_curve}
\end{figure}

The TRP is crucial to many problems in robotics, and has enabled better methods for nonconvex optimization, optimization on Reimannian Manifolds\cite{absil2007trust}, optimization with local approximations of dynamics \cite{suh2025dexterouscontactrichmanipulationcontact}, and even in highly efficient quadratic program solvers like HPIPM \cite{frisonHPIPMHighperformanceQuadratic2020} because of its
exploitation of problem structure for enhanced solving time.
From the KKT conditions, if $-Q^{-1}g$ is not in the interior of the feasible set, the trust region solution satisfies 
\[
(Q+\lambda I)x^*=-g,\quad \|x^*\|=\Delta
\]
for some scalar $\lambda\ge0$. The resulting curve of points respecting `stationarity conditions',
\begin{align}\label{eq:optimal_curve}
x(\lambda)=-(Q+\lambda I)^{-1}g
\end{align}
is the \emph{optimal curve}, spanning from the unconstrained minima $x(0)=-Q^{-1}g$ to the origin as $\lambda\to\infty$ as shown in Fig.~\ref{fig:TRP_opt_curve}. If the unconstrained optimum lies outside the trust region, finding the constrained optimum reduces to a one dimensional search over $\lambda$ to meet the norm bound. TRP solvers were explored in~\cite{gay1981computing,more1983computing}.

This search can be accelerated with an eigendecomposition $Q=V\Lambda V^\top$, which yields 
\[
(Q+\lambda I)^{-1}=V(\Lambda+\lambda I)^{-1}V^\top,
\]
allowing efficient evaluation of $\|(Q+\lambda I)^{-1}g\|^2 = V(\Lambda+\lambda I)^{-2}V^\top$ at each iteration. $(\Lambda+\lambda I)$ remains a diagonal matrix, making this inversion an elementwise operation.

\begin{figure} 
    \centering  \includegraphics[width=\columnwidth] {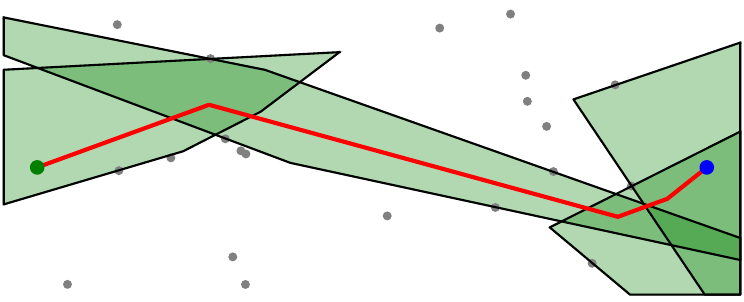}
    \caption{A decomposition of the space with polytopic corridors (in green). The polytopes are inflated about linear segments (red) to exclude collision points (grey).}\label{fig:IRIS_polyt_regions}
\end{figure}

\section{Nonconvex Obstacle Free Corridor Parameterizations}
\begin{figure}
    \centering  \includegraphics[width=\columnwidth,  trim=40 100 30 125, clip] {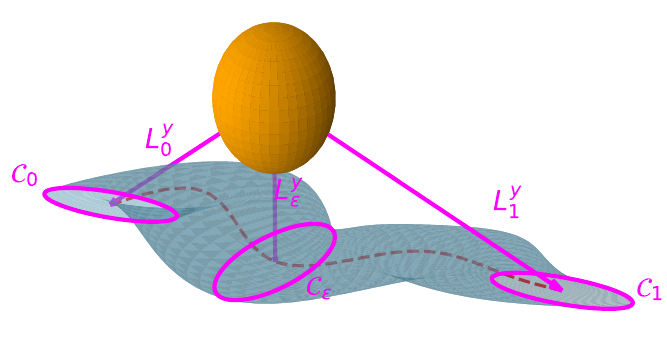}
    \caption{A corridor (blue) as a time-varying projection of a lifted hypersphere (orange). Example shown for $N_q = 2$ and $N_y = 3$.}\label{fig:proj_to_corridor}
\end{figure}

In this section we present a method for parameterizing trajectories in collision-free space using the ellipsoidal corridors described above. We represent the corridor as a set-valued trajectory $\mathcal{C}(\epsilon)$, shown as the blue region in Fig.~\ref{fig:proj_to_corridor}, where $0 \leq \epsilon \leq 1$ is the normalized ``time'' along the path ($\epsilon=0$ at the start and $\epsilon=1$ at the end).


In this work, we take an approach that defines trajectories in the configuration space by their offset from a reference trajectory. This reference trajectory serves as the centerline of our ellipsoidal corridor, and is denoted by $\tilde{q}_\epsilon$. Although $\tilde{q}_\epsilon$ can naturally be a feasible candidate of the trajectory of the system, it doesn't fully exploit the full obstacle-free configuration space, leading to conservative and often suboptimal trajectories. Hence, we would like to leverage a proxy of the obstacle-free space to optimize for the trajectory, which is treated as a variation $q_\epsilon$ of the centerline of the ellipsoidal corridor $\tilde{q}_\epsilon$. We can set our trajectory in the configuration space to be 
\begin{equation}
q_\epsilon = L_\epsilon y + \tilde{q}_\epsilon, 
\end{equation}
where $y\in\mathbb{R}^{N_y}$ is a vector of trajectory parameters our proposed method optimizes for and $L_\epsilon$ is a possibly non-polynomial function of $\epsilon$ that we will construct in the following Section.~\ref{sec:transformation_balls},  \ref{sec:corr_as_proj}, \ref{sec:proj_traj}. 

To ensure that resulting trajectories are collision free, we will restrict $L_\epsilon y$ to be an offset from $\tilde{q}_\epsilon$ such that the trajectory point is still within our collision-free ellipsoidal cross section. With this parameterization, optimizing a quadratic objective on the configuration or its derivatives is
\begin{align}
\min_{y} &\sum_{\epsilon_i \in [0,1]} \|Q_i \left(L^{(n)}_{\epsilon_i} y + \tilde{q}^{(n)}_{\epsilon_i}\right) + g_i\|
\\
\text{ s.t. }& \,\,\, q_{\epsilon_i} = L_{\epsilon_i} y + \tilde{q}_{\epsilon_i} \label{eq:opt_constriant_config} \\
&\,\,\, \tilde{q}_{\epsilon_i}\in \mathcal{C}_{\epsilon_i} \label{eq:opt_constriant_ball}
\end{align} 
where $q^{(n)}_\epsilon = L^{(n)}_{\epsilon} y + \tilde{q}^{(n)}_{\epsilon}$ is the $n$th derivative of the trajectory $q_\epsilon$. These contraints are complicated, and scale with the number of samples ($\epsilon_i$) used. We will instead define a set of valid parameter values $\mathcal{A}$ and parameterization $L_\epsilon$, such that
\[ 
\mathcal{C}_\epsilon = L_\epsilon \mathcal{A} + \tilde{q}_\epsilon.
\]

From this parameterization, any $y \in \mathcal{A}$ results in a configuration within collision free space, reflected in 
\[ 
q_\epsilon = L_\epsilon y + \tilde{q}_\epsilon \in \mathcal{C}_\epsilon \text{ if } y\in \mathcal{A},
\]
reducing the constraints in the optimization problem (\eqref{eq:opt_constriant_config}-\eqref{eq:opt_constriant_ball}) to simply $y \in \mathcal{A}$. Note that $\tilde{q}_\epsilon$ serves as the center of $\mathcal{C}_\epsilon$.

We now describe how we parameterize collision-free trajectories within the time-varying ellipsoidal corridors. We build up the parameterization in three steps. First, we show how a simple ball can be mapped to an ellipsoidal corridor cross-section. Next, we extend this to a lifted higher-dimensional space to introduce more trajectory diversity. Finally, we generalize to a Cartesian product of balls to capture boundary-hugging trajectories. This section thus moves from the simplest to the richest representations, culminating in our full parameterization.

Unlike~\cite{arrizabalagaDifferentiableCollisionFreeParametric2024a}, which restricts ellipsoidal cross-sections to degenerate planar slices perpendicular to a reference trajectory, our approach allows full-volume ellipsoids in the original configuration space. We define each cross-section by 
\begin{equation}\label{eq:corridor_definition_1}
\mathcal{C}_\epsilon=\{q\;|\;\|C_\epsilon(q-\tilde{q}_\epsilon)\|^2\leq1\},
\end{equation}
and use this definition to construct the mappings from $\mathcal{A}$ to $\mathcal{C}_\epsilon$. Here, $C_\epsilon$ represents the axes of the ellipsoid. In the following, we are going to present how we construct the set $\mathcal{A}$ and the corresponding mappings from $\mathcal{A}$ to $\mathcal{C}_\epsilon$.

\subsection{Corridors as transformations of balls ($\mathcal{A} =\mathcal{S}^{N_q}$)\label{sec:transformation_balls}}
We begin with the simplest case, mapping a unit ball $\mathcal{S}^{N_q}$ in configuration space into the time-varying ellipsoidal corridor cross sections.

Our cross section is defined by~\eqref{eq:corridor_definition_1} which is equivalent to
\begin{equation} \label{eq:ball_to_ellipse_Nq}
     \mathcal{C}_{\epsilon} = C_{\epsilon}^{-1} \mathcal{S}^{N_q} + \tilde{q}_\epsilon= \Big\{q = C_{\epsilon}^{-1} y + \tilde{q}_\epsilon \; | \; \|y\|^2 \leq 1\Big\}.
\end{equation} 
$\mathcal{S}^{N_q} = \{ y \in \mathbb{R}^{N_q}\; |\; \|y\|^2 \leq 1\}$ is the unit ball in $\mathbb{R}^{N_q}$. Equation~\eqref{eq:ball_to_ellipse_Nq} defines $\mathcal{C}_\epsilon$ as a varying transformation of a ball, $\mathcal{S}^{N_q}$, and defines the associated map as $L_\epsilon = C_\epsilon^{-1}$. 

Note that, in the above definition, columns of $C^{-1}_{\epsilon}$ define the ellipsoid axes. In the special case of a ball, $C^{-1}_{\epsilon}y = rI^{N_q}y$, effectively scales $y$ by $r$. 

let us find the trajectory that a single point in this $\epsilon$-varying ellipsoid traces. Given some point $y \in \mathcal{S}^{N_q}$, its corresponding trajectory is 
\[
q_\epsilon = C^{-1}_{\epsilon} y + \tilde{q}_{\epsilon}.
\]
Because $y$ is in $\mathcal{A} = \mathcal{S}^{N_q}$, the trajectory value $q_\epsilon$ remains in $\mathcal{C}_\epsilon$ for all $\epsilon$. Conversely, only a single unique trajectory ($y$) achieves configuration $q$ at $\epsilon$. The parameter for this trajectory is given by the inverse mapping,
\[
y = C_\epsilon (q - \tilde{q}_\epsilon).
\]

\subsection{Corridors as Projections of lifted balls ($\mathcal{A} = \mathcal{S}^{N_y}$)} \label{sec:corr_as_proj}
However, restricting ourselves to $y \in S^{N_q}$ limits the diversity of trajectories. With $\epsilon$ fixed, the above mapping between $q$ and $y$ is bijective, meaning each $q$ corresponds to exactly one trajectory ($y$). To overcome this, we lift to a higher-dimensional ball $S^{N_y}$ where $N_y > N_q$, as in Fig.~\ref{fig:proj_to_corridor}. We define our new map $L_\epsilon$ such that 
\[
\mathcal{C}_{\epsilon} = L_\epsilon\mathcal{S}^{N_y} + \tilde{q}_{\epsilon}, \quad N_y > N_q,
\]
where $L_\epsilon \in \mathbb{R}^{N_q \times N_y}$ now defines an injective map from $y$ to $q$. This creates an affine subspace 
\begin{equation} \label{eq:Y_e_q}
Y(\epsilon, q) = \{y \;|\; q = L_\epsilon y + l_{\epsilon}\}
\end{equation}
of parameter values that achieve $q$ at $\epsilon$. The intersection of this space and the feasible set ($Y(\epsilon,q)\cap\mathcal{S}^{N_y}$) reflects the volume of feasible trajectories in $\mathcal{A}=\mathcal{S}^{N_y}$ that pass through $q$ at $\epsilon$. By using this lifted space, we introduce a multitude of feasible trajectories that achieve a possible desired configuration at $\epsilon$. By constraining $y\in \mathcal{S}^{N_y}$ we guarantee the resulting trajectory stays inside the collision free corridor, while also obtaining a single norm constraint suitable for efficient Trust Region Subproblem solvers. 

\textbf{Building $\mathbf{L_\epsilon}$.} We will build $L_\epsilon$ by composing two transformation operations. The first is a projection from  $\mathcal{S}^{N_y}$ into $\mathcal{S}^{N_q}$. To project $\mathcal{S}^{N_y}$ into $\mathcal{S}^{N_q}$, we require a projection matrix $L^{\text{proj}}_\epsilon$. Combining $L^{\text{proj}}_\epsilon$ with $C_\epsilon^{-1}$ yields the ellipsoidal cross section of the corridor at $\epsilon$,
\begin{align} \label{eq:L_y_eps}
\mathcal{C}_\epsilon = L_\epsilon \mathcal{S}^{N_y} + \tilde q_\epsilon, \quad 
L_\epsilon=C_\epsilon^{-1}L^{\text{proj}}_\epsilon,
\end{align}
where
\[
q_\epsilon=L_\epsilon y + \tilde q_\epsilon,\quad q(\epsilon,y)\in \mathcal{C}_\epsilon \text{ if } \|y\|^2 \leq 1.
\]

We first show the conditions for the projection to $\mathcal{S}^{N_y}$ to be valid, then we attempt to find a valid $L^{\text{proj}}_\epsilon$ that respects the presented conditions for all $\epsilon$.

\begin{figure}
    \centering
    \includegraphics[width=\columnwidth,  trim=40 30 20 0, clip]{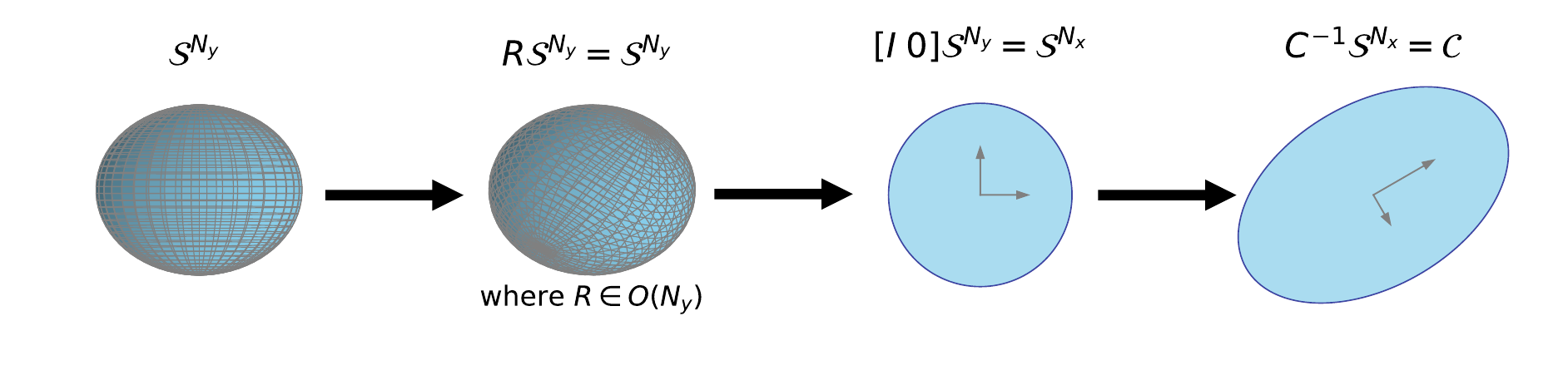}
    \caption{Projection of a lifted hypersphere to an ellipsoidal cross section: rotation in $\mathbb{R}^{N_y}$, projection to $\mathbb{R}^{N_q}$, and stretching to form $\mathcal{C}_\epsilon$.\label{fig:fig_flowchart}}
\end{figure}

\textbf{Validity of Projection Matrices.} We show here that any projection matrix $L^{\text{proj}}_\epsilon \in \mathbb{R}^{N_q \times N_y}$ with \textbf{orthonormal rows} defines a valid mapping from the unit ball in $\mathbb{R}^{N_y}$ to the unit ball in $\mathbb{R}^{N_q}$. Geometrically, applying this matrix to $\mathcal{S}^{N_y}$ is equivalent to rotating the ball in $\mathbb{R}^{N_y}$ and then taking its $N_q$-dimensional cross section, as illustrated in Fig.~\ref{fig:fig_flowchart}. 

These operations can be written as
\[
L^{\text{proj}}_\epsilon = [I_{N_q}\;0]R,
\quad \text{with } R \in O(N_y),
\]
which corresponds to selecting the first $N_q$ rows of an orthogonal matrix $R$, resulting in an $N_q \times N_y$ matrix with orthonormal rows. Conversely, given the top $N_q$ rows, we can always reconstruct an orthogonal $N_y \times N_y$ matrix $R$ (for example, via Gram–Schmidt completion).

\textbf{Constructing $\mathbf{L^{\text{proj}}_\epsilon}$.} 
We parameterize $L^{\text{proj}}_\epsilon$ as
\[
L^{\text{proj}}_\epsilon = I_{N_q} \otimes \bar B_\epsilon^\top,
\quad \bar B_\epsilon \in \mathbb{R}^{\frac{N_y}{N_q}},\;\|\bar B_\epsilon\|=1,
\]
where $\otimes$ is the Kronecker product. This guarantees each row has unit norm and rows are mutually orthogonal. We generate $\bar B_\epsilon$ as the normalized basis of a Bézier curve to obtain a smooth time-varying projection.

\subsection{Trajectories in the Lifted Space ($\mathcal{A} = (\mathcal{S}^{N_y})^{N_J}$)} \label{sec:proj_traj}
In spite of the ability of the above parameterization ($\mathcal{A} = \mathcal{S}^{N_y}, L_\epsilon = C^{-1}L^{\text{proj}}_\epsilon$) to describe multiple trajectories that pass through a given configuration, it does not have the ability to describe qualitatively ideal trajectories, motivating a new representation of $\mathcal{A}$ as a product of balls, $\mathcal{A} = \mathcal{S}^{N_y}\times\mathcal{S}^{N_y}\times...\mathcal{S}^{N_y}$.

\paragraph{Density of Describable Trajectories.}  

\begin{figure}
    \centering  \includegraphics[width=0.7\columnwidth,  trim=60 100 40 125, clip] {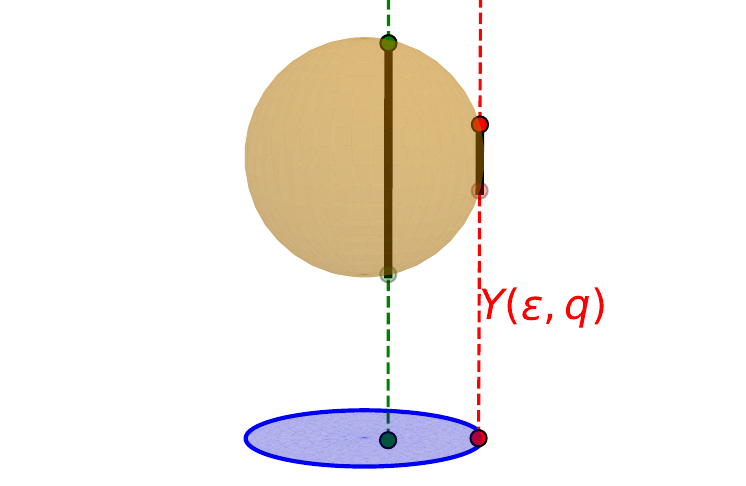}
    \caption{Feasible lifted parameters $y$ (black) projecting to two different $q$ values at $\epsilon$. Points closer to the corridor center admit a larger set of $y$, while boundary points admit fewer.}
\label{fig:projection_nullspace}
\end{figure}

To see how the previous parameterization limits feasible trajectories, consider $Y(\epsilon,q)$, the set of trajectories achieving configuration $q$ at $\epsilon$. From~\eqref{eq:Y_e_q}, a basic solution for $y \in Y(\epsilon, q)$ is $y^\ast=\left(L_\epsilon\right)^{\dagger}(q-\tilde q_\epsilon)$. All other solutions in this set are offset from $y^\ast$ by vectors in $\ker(L_\epsilon)$ so that \[Y(\epsilon,q)=y^\ast+z \quad \text{ where } y^\ast\perp z,\;\; z \in ker(L_\epsilon).\] The feasible set of trajectories in $Y(\epsilon, q)$, given by $Y(\epsilon, q) \cap \mathcal{S}^{N_y}$ yields a ball in $\ker(L_\epsilon)$ of radius $r(\epsilon,q)=\sqrt{1-\|y^\ast\|^2} = \sqrt{1-\|(L_\epsilon)^\dagger(q-\tilde q_\epsilon)\|}$. The radius
decreases with the Mahalanobis distance from the center, $\|(L_\epsilon)^\dagger(q-\tilde q_\epsilon)\|$, as visualized in Fig.~\ref{fig:projection_nullspace}. This shows that the previous parameterization naturally concentrates trajectories near the center of $\mathcal{C}_\epsilon$ while limiting exploration near its edges, which is undesirable when boundary-hugging paths are required.

\paragraph{Time-Varying Parameters}
To counter this concentration effect, we expand the parameter space to $\mathcal{A} = (\mathcal{S}^{N_y})^{N_J}$, the cartesian product of $N_J$ unit balls in $\mathbb{R}^{N_y}$.
\[
\mathbf{y} = [y_1, y_2, \ldots, y_{N_J}], \qquad y_j \in \mathcal{S}^{N_y} \;\forall j.
\]

We define an $\epsilon$-varying transformation from $(\mathcal{S}^{N_y})^{N_J}$ to $\mathbb{R}^{N_y}$ that traces a trajectory contained within the ball $\mathcal{S}^{N_y}$. This specifically results in $y_\epsilon = B_\epsilon\mathbf{y}$, where $y_\epsilon \in \mathcal{S}^{N_y}$, as visualized in Fig.~\ref{fig:proj_traj_to_corridor}. We can generate this mapping by taking an $\epsilon$-varying convex combination of $y_1, ..., y_{N_J}$ (e.g., a Bézier curve), which automatically stays inside the unit ball. This richer parameterization increases the amount of feasible trajectories that pass near the boundary of the corridor, capturing curved or boundary-hugging paths while maintaining the simple ball constraint structure that enables efficient Trust Region methods for optimization.

With $y(\epsilon) = B_\epsilon \mathbf{y}$ defining this trajectory, the projection from $(\mathcal{S}^{N_y})^{N_J}$ is given by 
\[
\mathcal{C}_\epsilon = L^\mathbf{y}_\epsilon(\mathcal{S}^{N_y})^{N_J} + \tilde{q}_\epsilon,\; L^\mathbf{y}_\epsilon =C^{-1}_\epsilon L^{\text{proj}}_\epsilon B_\epsilon,
\]
resulting in a trajectory in the configuration space given by 
\[
q(\epsilon, \mathbf{y}) = L^\mathbf{y}_\epsilon\mathbf{y} + \tilde{q}_\epsilon.
\]

\section{Solving Specialized QCQPs}

We have now introduced three main parameterizations of trajectories. In the first two, parameters are restricted to a single ball ($y \in \mathcal{S}^{N_q}$ or $y \in \mathcal{S}^{N_y}$). In the third, parameters restricted to a product of balls ($\mathbf{y}\in (\mathcal{S}^{N_y})^{N_J}$).

In the first two cases, our optimization problem is a Trust Region Problem. Based on this observation, let us develop specialized optimization methods that exploit the structure arising in the third case.

\textbf{The Orthogonal Trust Region Problem. }With the parameterization using a product of balls, the trajectory optimization takes the form 
\begin{align}\label{eq:orthtrp}
    \min_{\mathbf{y} = [y_1, ...y_{N_J}]^\top} \quad & \tfrac{1}{2} \mathbf{y}^\top Q \mathbf{y} + g^\top \mathbf{y} \\
    \text{s.t.}\quad & \|y_j\|^2 \leq 1\;\; \forall 1\le j \le N_J \label{eq:orthtrp:norm_constraints}
\end{align}

The feasible set of this problem is made of ball constraints enforced on orthogonal sets of axes. We dub this problem the `Orthogonal Trust Regions Problem'. 

\textbf{Separability.} An important observation is that the norm constraints \eqref{eq:orthtrp:norm_constraints} are separable, as each block variable $y_j$ has its own independent constraint $y_j \in \mathcal{S}^{N_y}$. 

Let us consider the optimization problem along only a cross section of the variables (a single block), with other variables fixed. The resulting subproblem is
\begin{align}\label{eq:orthtrp_block}
    \min_{y_i} \quad & \tfrac{1}{2} y_i^\top Q_{i,i} y_i 
    + \Big(g_i + \sum_{j \neq i} Q_{i,j}y_j\Big)^\top y_i \\
    \text{s.t.}\quad & \|y_i\|^2 \leq 1 \nonumber
\end{align}
which is a traditional trust region problem. Note here that $Q_{i,j}$ is a submatrix of $Q$ whose columns correspond to the block $y_i$ and rows to $y_j$. Since $Q$ is symmetric, $Q_{i,j}^\top = Q_{j,i}$ 

Due to the separability of the problem, the feasible set for $y_i$ is constant, regardless of values for $y_{j \neq i}$. This is a desirable structure, as the iteration on the solution for each block can be done by solving (or approximately solving) a TRP.

This motivates a block coordinate descent (BCD) strategy for solving \eqref{eq:orthtrp}. In each iteration we cycle through the blocks, updating $y_i$ by solving \eqref{eq:orthtrp_block}. Rather than solving each subproblem to exact optimality, we may perform only a single dual update step for the corresponding multiplier $\lambda_i$, followed by the primal update $y_i$. This yields an inexact BCD strategy that exploits the Trust Region like structure to make each step computationally inexpensive.

\begin{figure} 
    \centering  \includegraphics[width=\columnwidth,  trim=130 10 120 10, clip] {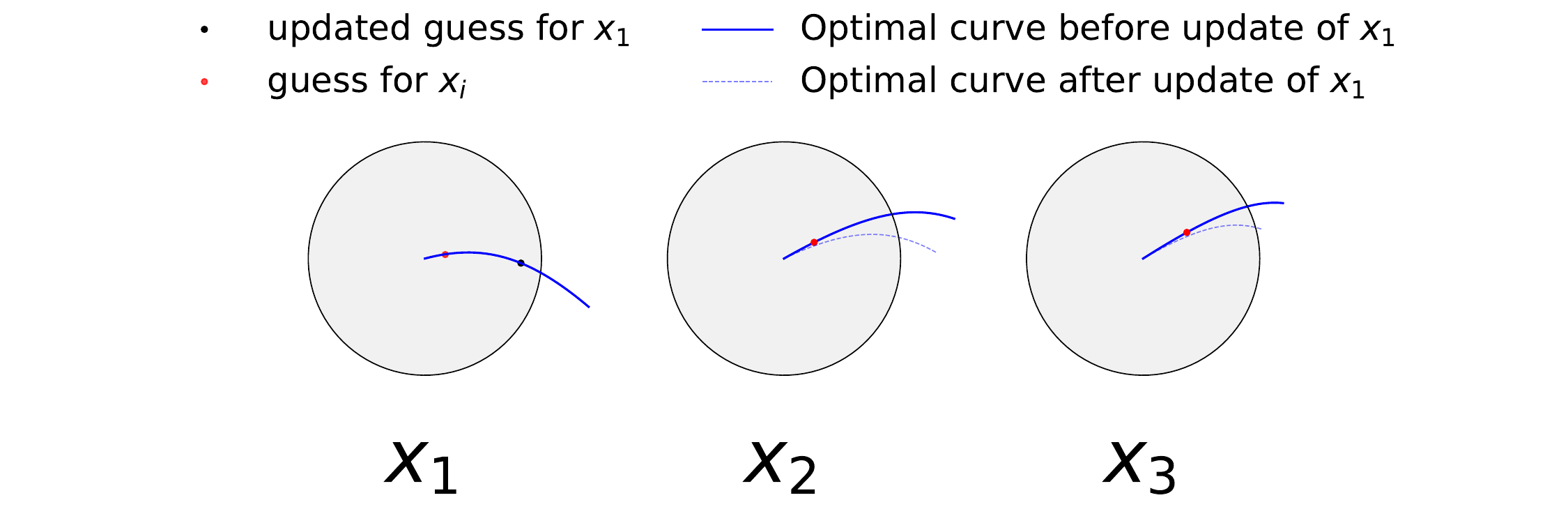}
    \caption{A visualization of the Orth-TRP problem as $N_J=3$ interconnected Trust Region Problems. As the guess for $x_1$ is updated, the optimal curves for $x_2$ and $x_3$ are influenced, changing the optimal solution for those parallel problems.}\label{fig:orthtrp}
\end{figure}

\noindent\textbf{Geometric intuition.}  
Let us look at the subproblem in each separable direction. From \eqref{eq:optimal_curve} and \eqref{eq:orthtrp_block}, the optimal curve in direction $i$ is given by 
\[
y_i(\lambda_i) = -(Q_{i,i} + \lambda_i I)^{-1}\left(g_i +\sum_{1\leq j \leq N_J} Q_{i,j} y_j\right)
\]
while the feasible set is constant. 

We can view the problem as $N_J$ interconnected trust region problems. As we update the guess for the optimal value in some direction, $y_j$, the objective of the other $N_J - 1$ subproblems (and thus the relevant optimal curve) is changed slightly, as shown in Fig.~\ref{fig:orthtrp}. 

\textbf{Algorithm.} We can implement the proposed solver in Algorithm~\ref{alg:bcd-trp-simple}, using a Moré--Sorensen style trust region step for each block. Only
$c_i = g_i + \sum_{j\neq i} Q_{ij}\,y_j$
changes between iterations, while $Q_{ii}$ remains constant; its eigendecomposition can be precomputed and reused. Each \textsc{TRP\_Step} updates the dual variable, $\lambda_i$ for the direction $i$.

\begin{algorithm}[t]
\caption{Orth-TRP solver with Moré--Sorensen block updates}
\label{alg:bcd-trp-simple}
\begin{algorithmic}[1]
\Require $Q_{ij}\in\mathbb{R}^{N_y\times N_y}$ with $Q_{ii}\succ 0$, vectors $g_i\in\mathbb{R}^{N_y}$, $y_i\in\mathcal{S}^{N_y}$ for $i=1,\dots,N_J$.
\State Precompute eigendecomposition of each $Q_{ii}$.
\State Initialize $y_i^{(0)}\gets 0$; $\lambda_i^{(0)} \gets \infty$ set $k\gets 0$.
\Repeat
  \For{$i=1$ to $N_J$} \Comment{Gauss--Seidel sweep}
    \State $c_i^{(k)} \gets g_i + \sum_{j\neq i} Q_{ij}\,y_j^{(k+\mathbb{1}[j<i])}$
    \State $y_i^{(k+1)} , \lambda_i^{(k+1)} \gets \text{TRP\_Step}_{\text{Moré--Sorensen}}\big(Q_{ii},c_i^{(k)}, \lambda_i^{(k)}\big)$
  \EndFor
  \State $k\gets k+1$
\Until{converged}
\State \textbf{return} $[y_1^{(k)};\dots;y_{N_J}^{(k)}]$
\end{algorithmic}
\end{algorithm}

\section{Results}

While our approach is broadly applicable to collision free planning, we evaluate our method on a quadrotor trajectory optimization problem designed to highlight how solver performance scales with trajectory complexity.

\begin{figure}[t]
    \centering
    \begin{subfigure}[b]{0.55\columnwidth}
        \centering
        \includegraphics[width=\linewidth]{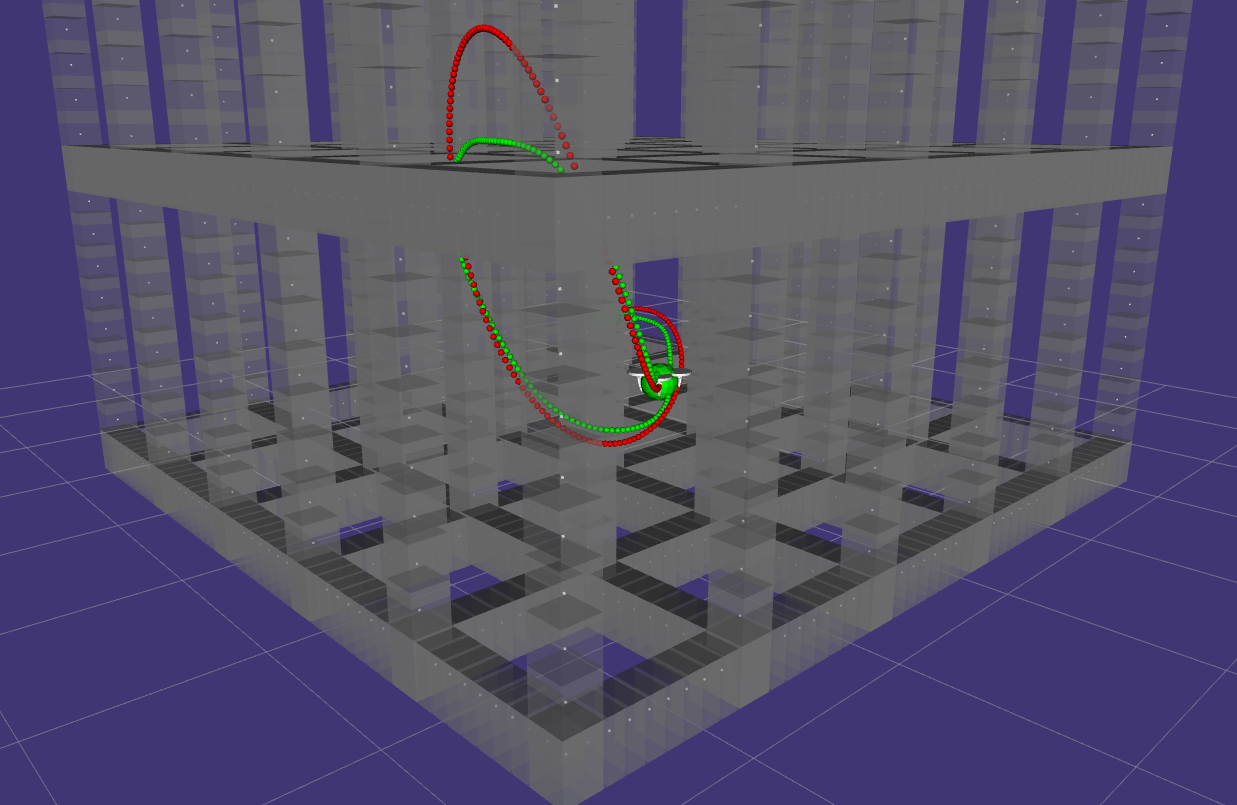}
        \caption{An example solution trajectory (in green) using ellipsoidal corridors in our grid of rooms. This is contrasted with the smooth reference trajectory (red). Note that the solution is able to `hug' the boundary of the corridor, and pass near obstacles. }
        \label{fig:solution_traj_example}
    \end{subfigure}
    \hfill
    \begin{subfigure}[b]{0.38\columnwidth}
        \centering
        \includegraphics[width=\linewidth]{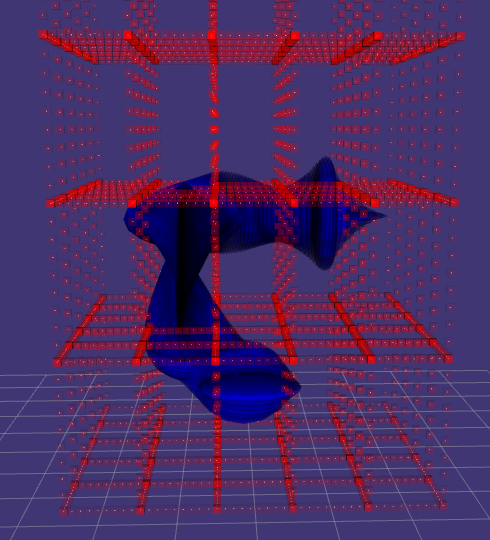}
        \caption{An example ellipsoidal corridor in the collision free space of our example environment. Trajectories are restricted to be within this corridor.}\label{fig:ballal_corridor_example}
    \end{subfigure}
    \label{fig:combined_figure}
\end{figure}

\textbf{Experimental Setup.} 
We evaluate our approach on a quadrotor trajectory optimization task to show how solver performance scales with trajectory complexity. The environment is a 3D grid of adjacent rectangular ``rooms,'' with collision points sampled along shared edges (Fig.~\ref{fig:solution_traj_example}). Each experiment attempts to generate a trajectory moving from one room to the next, minimizing acceleration.
We measure two main outcomes. The first is solver time, contrasting a large polytope based QP (solved with OSQP a state of the art QP solver) with our Orth-TRP solver. The second is trajectory quality, quantified by the relative improvement in average acceleration over a naively generated trajectory. By varying the number of waypoints and homotopy classes, we analyze the performance.

\textbf{Initial Trajectories and Homotopy Classes.} 
We generate a linearly interpolated initial guess for each trial, where a trajectory visiting rooms ``A--B--C'' interpolates through the shared walls between A and B, then between B and C. Each initial guess defines a distinct homotopy class. The number of waypoints, varied from 11 to 30, controls trajectory curvature and acts as a proxy for topological complexity.

\textbf{Smooth Reference Trajectory and Corridors.} 
We fit a 15 control point Bézier curve to the linearly interpolated trajectory to generate a smooth reference trajectory with zero initial and final velocity. Around each sampled point we solve a linear program to fit a large, axis aligned ellipsoid. Interpolating between the parameters of these ellipsoids helps generate the time-varying corridor $\mathcal{C}$.

\textbf{Decision Variables.} 
Our planner uses $N_y=6$ and $N_J=14$ orthogonal blocks, fixing the first and last two blocks to enforce zero velocity at the boundaries. The remaining variables capture free directions within the ellipsoidal corridor. For the polytope baseline, we enforce continuity of position, velocity, and acceleration, with 5 control points (resulting in $5N_q = 15$ variables) per polytope; with $N$ waypoints, this yields $15(N-1)$ variables in total.

\textbf{Baseline Methods.} 
\textit{Polytopic Corridor Method~\cite{liuPlanningDynamicallyFeasible2017}}: We implemented the planner in \cite{liuPlanningDynamicallyFeasible2017}, which builds convex polytopes along the initial trajectory, and uses those to generate a polytopic corridor. We allocate time across polytopes by computing the `start time' for each segment in the polytope with
\begin{align}
s_n^\ast = \frac{\text{arc length for segments before }n}{\text{total arc length}} \nonumber
\\
\text{start time}^n=\text{smoothstep}^{-1}(s_n^\ast)\times\text{horizon}\nonumber
\end{align}
which proportionally enlarges time for longer segments. Inverting the `smoothstep' function allows this approach to allocate larger amounts of time for the start and end of the trajectory to account for acceleration from (or deceleration to) rest. Although this is a heuristic, it produces smoother timing than uniform allocation.

\textit{Bubble Planning}: We also reproduced the ``Bubble Planning'' method, which uses overlapping spheres along the path and solves a QCQP with NLOPT. Waypoints are chosen from the intersections of these spheres, and the solution is formed by interpolating between them. As the waypoint count increased (necessary for the complex trajectories considered here), the interpolated paths exhibited unpredictable, oscillatory behavior, and the resulting trajectories became infeasible in all trials with more than 13 waypoints.

\subsection{Solving Time vs. Trajectory Complexity}
In contrast to existing corridor-based planners, the runtime and problem size of our method does not scale with regards to the solution complexity or the geometric complexity of the environment.

\begin{figure}
    \centering  
    \includegraphics[width=\columnwidth]{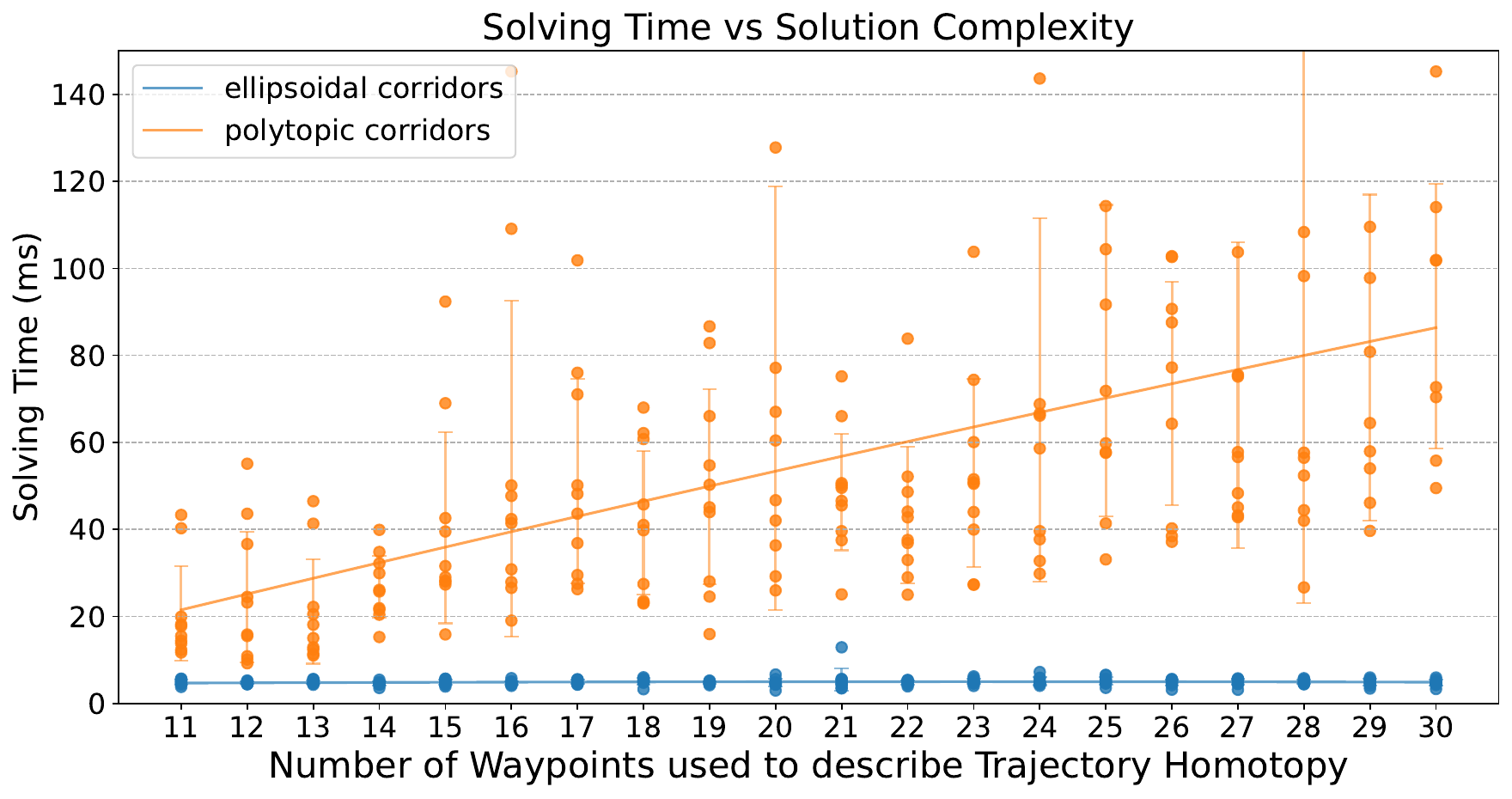}
    \caption{Solver runtimes versus trajectory complexity. Polytopic corridors (blue) become slower as waypoint count increases, while ellipsoidal corridors with Orth-TRP (orange) maintain nearly constant runtime across all tested complexities.}
    \label{fig:solving_times}
\end{figure}

Fig.~\ref{fig:solving_times} directly demonstrates this. Because the lifted, ellipsoidal parameterization and the Orth-TRP solver decouple problem size from geometric complexity, solver runtime stays nearly constant even as the number of waypoints increases dramatically. At low complexity ($N=11$ waypoints), our method is already faster than the polytope-based QP. As the path becomes more complex, this gap widens. By contrast, the solving time of the polytope-based solver scales with the number of waypoints, reflecting the burden of utilizing increasingly numerous convex regions to describe environments with complicated geometries. This result confirms the core contribution of our approach; by lifting trajectories into a product-of-balls representation and solving an Orth-TRP, we achieve substantial computational gains over traditional planners.

The Orth-TRP solver itself consists of two stages. The first is a parallelizable eigendecomposition of $N_J$ $N_y \times N_y$ matrices (which dominates computation time). The second is an inexpensive per-block trust-region updates with matrix multiplication and elementwie inversion operations. In our experiments, eigendecomposition was about $8\times$ more expensive than the iterations, highlighting how the solver’s structure concentrates computational effort in a single, parallelizable step. This makes it feasible to plan smooth, dynamically feasible trajectories in environments far more complex than those tractable with polytopic corridors.

\subsection{Comparison of Solution Optimality}

\begin{figure}
    \centering
    \includegraphics[width=\columnwidth]{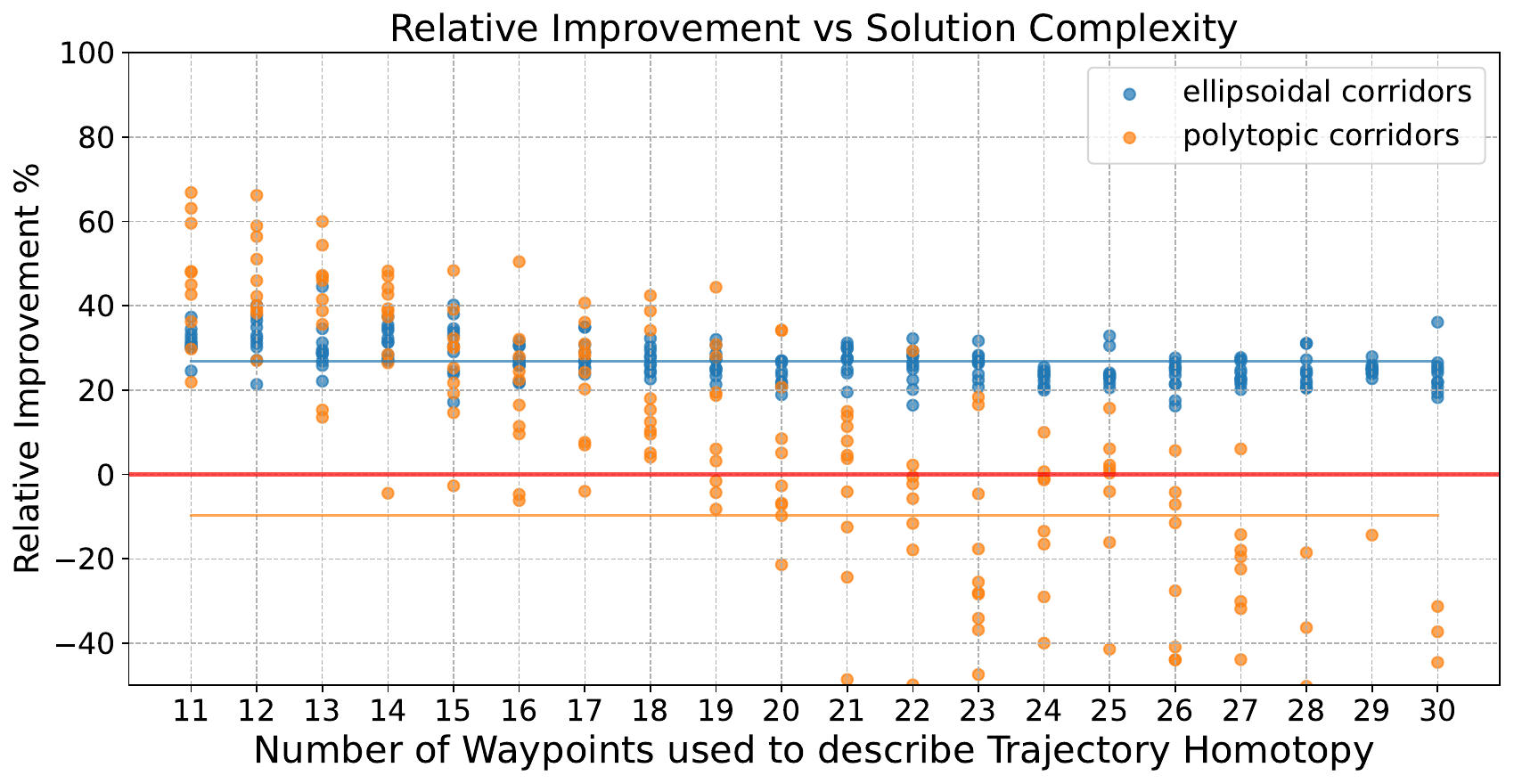}
    \caption{Average acceleration of optimized trajectories versus a naive reference. Orth-TRP trajectories (blue) remain smoother across all complexities than polytopic corridors (orange), with the advantage increasing as the path becomes more complex. The average cost improvement across all trials (horizontal lines) was negative for the polytopic corridors.}
    \label{fig:solving_accel}
\end{figure}

\begin{figure}
    \centering
    \includegraphics[width=\columnwidth]{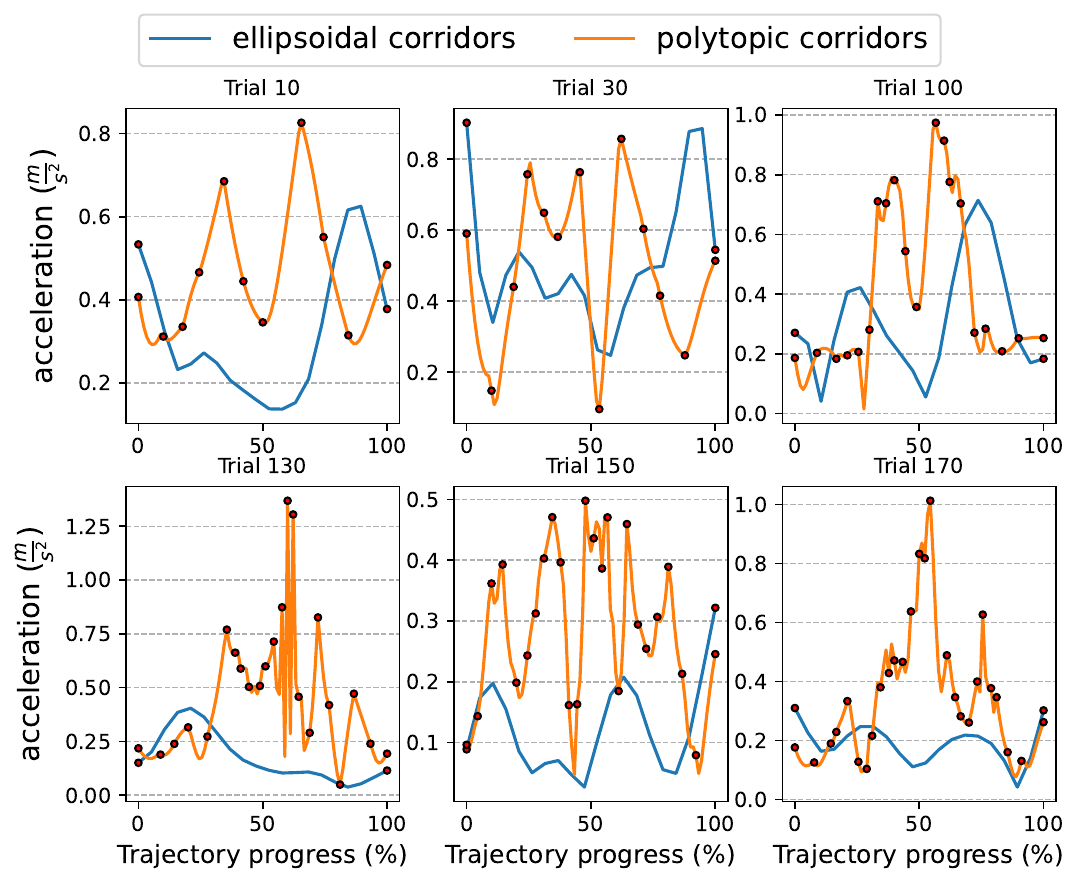}
    \caption{Representative acceleration profiles showing how Orth-TRP trajectories (blue) avoid the sharp peaks visible at discrete switching points (red dots) in polytopic corridors (orange).}
    \label{fig:solving_accel_profiles}
\end{figure}

Our method avoids the sharp acceleration peaks characteristic of other corridor-based approaches that assign fixed time intervals to each segment, as illustrated in Fig.~\ref{fig:solving_accel}, which highlights how our approach resolves this issue. Across all tested complexities, Orth-TRP yields consistently smoother, more dynamically feasible trajectories compared to both the naive smooth reference and the polytope-based baseline. Representative acceleration profiles in Fig.~\ref{fig:solving_accel_profiles} show that our method avoids the pronounced peaks that occur at discrete switching points (marked by red dots) in polytopic corridors.

This improvement stems directly from our key contribution, replacing discrete, time-allocated polytopes with a continuously deforming ellipsoidal corridor parameterization. By allowing the trajectory to evolve inside smoothly varying ellipsoidal cross sections and lifting to a Cartesian product of balls, we create a feasible set that naturally accommodates boundary-hugging and curved paths without forcing discontinuous accelerations. This explains why our method achieves lower acceleration costs and better scaling with increasing geometric complexity.

Overall, these results demonstrate that Orth-TRP not only solves the scalability problem but also mitigates the time-allocation and discontinuity issues that have historically limited corridor-based planners. By combining an efficient, block-separable QCQP solver with a rich parameterization of feasible paths, our approach produces trajectories that are both faster to compute and higher in quality in highly complicated environments.

\section{Conclusions}
We have presented a new framework for collision free trajectory optimization that treats safe corridors as projections of orthogonal trust regions. By lifting trajectories into a higher dimensional parameter space and representing feasible sets as Cartesian products of balls, we project this structure into time-varying ellipsoidal corridors. This decouples geometric complexity from problem size while maintaining a convex and scalable formulation. Unlike traditional corridor methods, which require difficult time allocation across convex regions, our approach allows trajectories to evolve smoothly inside ellipsoidal corridors without assigning fixed time intervals. Building on this representation, we formulated the Orthogonal Trust Regions Problem (Orth-TRP), a specialized QCQP whose separable nature results in optimization over interconnected Trust Region subproblems. Our Orth-TRP solver exploits this structure by reusing per-block eigendecompositions and solving each trust region dual via a simple scalar search, producing a parallelizable solver.

Future work will extend this parameterization to higher-DOF systems, cooperative manipulation, and receding-horizon planning with inexact Newton updates over the eigendecomposition, enabling a sequential updating method for similar problems. In summary, this work demonstrates a principled, convex, and scalable approach to planning highly curved trajectories in complex environments, opening new possibilities for advanced robotic motion planning.

\bibliographystyle{IEEEtran}
\bibliography{main}

\begin{thebibliography}{10}
\providecommand{\url}[1]{#1}
\csname url@samestyle\endcsname
\providecommand{\newblock}{\relax}
\providecommand{\bibinfo}[2]{#2}
\providecommand{\BIBentrySTDinterwordspacing}{\spaceskip=0pt\relax}
\providecommand{\BIBentryALTinterwordstretchfactor}{4}
\providecommand{\BIBentryALTinterwordspacing}{\spaceskip=\fontdimen2\font plus
\BIBentryALTinterwordstretchfactor\fontdimen3\font minus \fontdimen4\font\relax}
\providecommand{\BIBforeignlanguage}[2]{{%
\expandafter\ifx\csname l@#1\endcsname\relax
\typeout{** WARNING: IEEEtran.bst: No hyphenation pattern has been}%
\typeout{** loaded for the language `#1'. Using the pattern for}%
\typeout{** the default language instead.}%
\else
\language=\csname l@#1\endcsname
\fi
#2}}
\providecommand{\BIBdecl}{\relax}
\BIBdecl

\bibitem{marcucci2023gcs}
T.~Marcucci, M.~Petersen, D.~von Wrangel, and R.~Tedrake, ``Motion planning around obstacles with convex optimization,'' \emph{Science Robotics}, vol.~8, no.~84, p. eadf7843, 2023.

\bibitem{PallarLiLoianno2025_CBF_Manip_ICRA}
A.~Pallar, G.~Li, and G.~Loianno, ``Optimal trajectory planning for cooperative manipulation with multiple quadrotors using control barrier functions,'' in \emph{IEEE International Conference on Robotics and Automation (ICRA)}, 2025.

\bibitem{MaoEtAl2021_PersepPerching_IROS}
J.~Mao, G.~Li, S.~Nogar, C.~Kroninger, and G.~Loianno, ``Aggressive visual perching with quadrotors on inclined surfaces,'' in \emph{IEEE/RSJ Int. Conf. on Intelligent Robots and Systems (IROS)}, 2021.

\bibitem{LaValle1998RRT}
S.~M. LaValle, ``Rapidly-exploring random trees: A new tool for path planning,'' Department of Computer Science, Iowa State University, Ames, IA, USA, Tech. Rep. TR 98-11, oct 1998.

\bibitem{karamanSamplingbasedAlgorithmsOptimal2011}
S.~Karaman and E.~Frazzoli, ``Sampling-based algorithms for optimal motion planning,'' \emph{The International Journal of Robotics Research}, vol.~30, no.~7, pp. 846--894, Jun. 2011.

\bibitem{kavraki1996probabilistic}
L.~E. Kavraki, P.~Svestka, J.-C. Latombe, and M.~H. Overmars, ``Probabilistic roadmaps for path planning in high-dimensional configuration spaces,'' \emph{{IEEE} Transactions on Robotics and Automation}, vol.~12, no.~4, pp. 566--580, 1996.

\bibitem{TracyHowellManchester2023}
K.~Tracy, T.~A. Howell, and Z.~Manchester, ``Differentiable collision detection for a set of convex primitives,'' in \emph{IEEE International Conference on Robotics and Automation (ICRA)}, 2023, pp. 3663--3670.

\bibitem{foehnTimeoptimalPlanningQuadrotor2021}
P.~Foehn, A.~Romero, and D.~Scaramuzza, ``Time-optimal planning for quadrotor waypoint flight,'' \emph{Science Robotics}, vol.~6, no.~56, p. eabh1221, Jul. 2021.

\bibitem{LiLoianno2023_NMPC_CoopTransp_IROS}
G.~Li and G.~Loianno, ``Nonlinear model predictive control for cooperative transportation and manipulation of cable suspended payloads with multiple quadrotors,'' in \emph{IEEE/RSJ International Conference on Intelligent Robots and Systems (IROS)}, 2023.

\bibitem{thirugnanamSafetyCriticalControlPlanning2022}
A.~Thirugnanam, J.~Zeng, and K.~Sreenath, ``Safety-{{Critical Control}} and {{Planning}} for {{Obstacle Avoidance}} between {{Polytopes}} with {{Control Barrier Functions}},'' in \emph{2022 {{International Conference}} on {{Robotics}} and {{Automation}} ({{ICRA}})}, May 2022, pp. 286--292.

\bibitem{GoarinLiSavioloLoianno2025_DNMPC_Safety_ICRA}
M.~Goarin, G.~Li, A.~Saviolo, and G.~Loianno, ``Decentralized nonlinear model predictive control for safe collision avoidance in quadrotor teams with limited detection range,'' in \emph{IEEE International Conference on Robotics and Automation (ICRA)}, 2025.

\bibitem{JaitlyMERL2025}
A.~Jaitly, D.~Jha, K.~Ota, and Y.~Shirai, ``Analytic conditions for differentiable collision detection in trajectory optimization.''

\bibitem{deitsComputingLargeConvex2015}
R.~Deits and R.~Tedrake, ``Computing {{Large Convex Regions}} of {{Obstacle-Free Space Through Semidefinite Programming}},'' in \emph{Algorithmic {{Foundations}} of {{Robotics XI}}}, H.~L. Akin, N.~M. Amato, V.~Isler, and A.~F. {van der Stappen}, Eds.\hskip 1em plus 0.5em minus 0.4em\relax Cham: Springer International Publishing, 2015, vol. 107, pp. 109--124.

\bibitem{PetersenTedrake2023IRISNP}
M.~Petersen and R.~Tedrake, ``Growing convex collision-free regions in configuration space using nonlinear programming,'' \emph{arXiv preprint arXiv:2303.14737}, 2023, preprint.

\bibitem{liuPlanningDynamicallyFeasible2017}
S.~Liu, M.~Watterson, K.~Mohta, K.~Sun, S.~Bhattacharya, C.~J. Taylor, and V.~Kumar, ``Planning {{Dynamically Feasible Trajectories}} for {{Quadrotors Using Safe Flight Corridors}} in 3-{{D Complex Environments}},'' \emph{IEEE Robotics and Automation Letters}, vol.~2, no.~3, pp. 1688--1695, Jul. 2017.

\bibitem{renBubblePlannerPlanning2022}
Y.~Ren, F.~Zhu, W.~Liu, Z.~Wang, Y.~Lin, F.~Gao, and F.~Zhang, ``Bubble {{Planner}}: {{Planning High-speed Smooth Quadrotor Trajectories}} using {{Receding Corridors}},'' in \emph{2022 {{IEEE}}/{{RSJ International Conference}} on {{Intelligent Robots}} and {{Systems}} ({{IROS}})}, Oct. 2022, pp. 6332--6339.

\bibitem{arrizabalagaDifferentiableCollisionFreeParametric2024a}
J.~Arrizabalaga, Z.~Manchester, and M.~Ryll, ``Differentiable {{Collision-Free Parametric Corridors}},'' in \emph{{{IEEE}}/{{RSJ International Conference}} on {{Intelligent Robots}} and {{Systems}} ({{IROS}})}, Oct. 2024, pp. 1839--1846.

\bibitem{StellatoBanjacGoulartBemporadBoyd2020_OSQP}
B.~Stellato, G.~Banjac, P.~Goulart, A.~Bemporad, and S.~Boyd, ``{OSQP}: an operator splitting solver for quadratic programs,'' \emph{Mathematical Programming Computation}, vol.~12, no.~4, pp. 637--672, 2020.

\bibitem{absil2007trust}
P.-A. Absil, C.~G. Baker, and K.~A. Gallivan, ``Trust-region methods on riemannian manifolds,'' \emph{Foundations of Computational Mathematics}, vol.~7, no.~3, pp. 303--330, 2007.

\bibitem{suh2025dexterouscontactrichmanipulationcontact}
\BIBentryALTinterwordspacing
H.~J.~T. Suh, T.~Pang, T.~Zhao, and R.~Tedrake, ``Dexterous contact-rich manipulation via the contact trust region,'' 2025. [Online]. Available: \url{https://arxiv.org/abs/2505.02291}
\BIBentrySTDinterwordspacing

\bibitem{frisonHPIPMHighperformanceQuadratic2020}
G.~Frison and M.~Diehl, ``{{HPIPM}}: A high-performance quadratic programming framework for model predictive control,'' \emph{IFAC-PapersOnLine}, vol.~53, no.~2, pp. 6563--6569, 2020.

\bibitem{gay1981computing}
D.~M. Gay, ``{Computing optimal steps for trust region steps},'' Bell Laboratories, Murray Hill, NJ, USA, Tech. Rep. no. 6, 1981, this is a technical memorandum that was influential in the development of trust-region methods.

\bibitem{more1983computing}
J.~J. Mor{\'e} and D.~C. Sorensen, ``Computing a trust region step,'' \emph{SIAM Journal on Scientific and Statistical Computing}, vol.~4, no.~3, pp. 553--572, 1983.

\end{thebibliography}

\end{document}